\documentclass{article}
\usepackage{ijcai24}

\usepackage{times}
\usepackage{soul}
\usepackage{url}
\usepackage[hidelinks]{hyperref}
\usepackage[utf8]{inputenc}
\usepackage[small]{caption}
\usepackage{graphicx}
\usepackage{amsmath}
\usepackage{amsthm}
\usepackage{booktabs}
\usepackage{algorithm}
\usepackage{algorithmic}
\usepackage[switch]{lineno}
\usepackage{multirow}
\usepackage{subfig}
\usepackage{amsfonts}

\title{Rethinking Closed-loop Planning Framework for Imitation-based Model Integrating Prediction and Planning}

\author{
Jiayu Guo$^1$
\and
Mingyue Feng$^2$\and
Pengfei Zhu$^2$\and
Chengjun Li$^2$\And
Jian Pu$^1$\footnote{Corresponding Author:Jian Pu}
\affiliations
$^1$Institute of Science and Technology for Brain-Inspired Intelligence, Fudan University\\
$^2$Mogo Auto Intelligence and Telematics Information Technology Company Ltd.
\emails
\{jyguo20,jianpu\}@fudan.edu.cn,
\{fengmingyue, zhupengfei, james\}@zhidaoauto.com
}
\begin{document}

\maketitle

\begin{abstract}
In recent years, the integration of prediction and planning through neural networks has received substantial attention. Despite extensive studies on it, there is a noticeable gap in understanding the operation of such models within a closed-loop planning setting. To bridge this gap, we propose a novel closed-loop planning framework compatible with neural networks engaged in joint prediction and planning. The framework contains two running modes, namely planning and safety monitoring, wherein the neural network performs Motion Prediction and Planning (MPP) and Conditional Motion Prediction (CMP) correspondingly without altering architecture. We evaluate the efficacy of our framework using the nuPlan dataset and its simulator, conducting closed-loop experiments across diverse scenarios. The results demonstrate that the proposed framework ensures the feasibility and local stability of the planning process while maintaining safety with CMP safety monitoring. Compared to other learning-based methods, our approach achieves substantial improvement.
\end{abstract}

\section{Introduction}

In the field of autonomous driving, considerable attention has been devoted to the integration of prediction and planning using neural networks~\cite{rhinehart2019precog,huang2023gameformer,huang2023differentiable}. However, scant emphasis has been placed on understanding how such models operate to reduce accumulative errors within a closed-loop setting. 

Problems stemming from accumulative error in closed-loop planning have been identified in previous investigations of rule-based planner, such as lateral offset overshoot and maneuverability decrease~\cite{werling2010optimal}.
These issues manifest as excessively large or small curves in the controlled path during turns or lane change.  
Werling~\textit{et al}~\shortcite{werling2010optimal} points out that this deterioration results from a violation of Bellman’s principle of optimality for consecutive plans, while 
Jian~\textit{et al}~\shortcite{jian2020multi} explains it for the untimely replanning can disrupt the $G^{2}$ continuity of the global planning. 
Wu \textit{et al}~\shortcite{wu2022trajectory} observes similar issues in neural planner and propose trajectory planning branch to guide vehicle driving, but these concerns have yet to be addressed within the context of a closed-loop framework.

Typically, neural planners within closed-loop settings operate at a fixed high frequency, a characteristic stemming from the initial design principles of End-to-End (E2E) neural planners. 
These models, taking raw data as input and directly outputting control commands~\cite{codevilla2019exploring,prakash2021multi,xiao2020multimodal}, implicitly hinge their safety on learning to output commands that are deemed safe for the current environment. Thus, maintaining a fixed high execution frequency becomes imperative for safety assurance, especially in dynamic environments. However, as neural planners evolve to incorporate more interpretability, the neural planners become supervised by multi-tasks and a diverse range of intermediate outputs can be utilized, including planned trajectories of ego vehicle~\cite{wu2022trajectory}, prediction of surrounding vehicles~\cite{chen2022learning} and occupancy forecasting~\cite{sadat2020perceive}. 
In the presence of explicitly expressed intermediate output, the planning framework can be more flexible, adapting accordingly to reduce cumulative errors during closed-loop operation.


Inspired from~\cite{jian2020multi}, we establish a stand-alone safety monitor in closed-loop planning. Our framework not only obviates the necessity for the model to plan at a high frequency to ensure safety, but also can novely leverage Conditional Motion Prediction (CMP) to characterize the ego vehicle as a leader in agent interactions when evaluating environmental risks. This prevents the ego vehicle from adopting an overly conservative driving manner during travel. Our neural model, equiped with a subset of the non-neural modules, effectively fulfills the requirements of this framework without necessitating alterations to the network architecture across different operations. 

Our contributions can be summarized into three parts. First, we design an imitation-based modular graph network integrating prediction and planning, showcasing exceptional data efficiency through agent-centric representation. Then, a closed-loop planning framework fully compatible with that neural planner is introduced, which novelly considers the utilization of CMP for safety monitoring and performs adaptive scheduling. Finally, through extensive closed-loop simulations across diverse scenarios, we illustrate the advantages of the combination of our imitation-based integration model and the proposed planning framework.
\section{Related Work}
    
\subsection{Neural Planning with Prediction}
The autonomous planner landscape can be categorized into modular and End-to-End (E2E) methods based on their pipeline type. E2E neural planners derive actions directly from raw sensors~\cite{codevilla2019exploring,jia2023think}, while modular planners necessitate explicit output from upstream perception modules~\cite{bansal2018chauffeurnet,scheel2022urban,huang2023differentiable}.

Prediction as auxiliary task can implicitly facilitate learning to plan. Modular planners offer flexibility in modeling interactions, and naturally treat prediction and planning tasks as joint multi-agent prediction on ego and surrounding vehicles~\cite{rhinehart2019precog}. Relatively, Chen and Kr{\"a}henb{\"u}hl~\shortcite{chen2022learning} have highlighted that leveraging predictions of nearby agents as auxiliary tasks can improve sample efficiency for ego planning. Additionally, GameFormer~\cite{huang2023gameformer} directly models k-level interactive games among all agents through a k-layer transformer decoder, giving highly interactive prediction and planning. Conversely, some E2E approaches opt not to construct explicit instance-level features for each traffic participant from ego-centric sensors~\cite{codevilla2019exploring,prakash2021multi,xiao2020multimodal,wu2022trajectory}, aiming to reduce compounding errors or information bottlenecks. While this approach limits the ability to model social interaction and predict dynamics, it can facilitate learning to interact with other agents by implicit regularization from multi-task learning, such as occupancy forecasting~\cite{sadat2020perceive}. 
To enhance interaction modeling, certain E2E methods extract small agent-centric crops over intermediate feature maps to create their own representations~\cite{cui2021lookout}.

Explicit safety guarantees often entail the use of hand-crafted cost incorporated with prediction. DSDnet~\cite{zeng2020dsdnet} and P3~\cite{sadat2020perceive}, for instance, employ a collision penalty associated with occupancy forecasting or agent prediction in the planning cost for candidate selection. UniAD~\cite{hu2023planning} prevents planned trajectories from colliding by penalizing distance with predicted occupancy during post-optimization. To simplify the weight-tuning process, Diffstacks~\cite{karkus2023diffstack} and DIPP~\cite{huang2023differentiable} employ differentiable nonlinear optimizers, learning weights of the cost function from data.


\subsection{Conditional Motion Prediction}
\label{subsection:conditional motion prediction}
Motion prediction depending on certain predetermined agent future is known as conditional motion prediction (CMP). CMP can help improve model prediction accuracy in the sense of $L_{2}$ norm between prediction and ground truth~\cite{rhinehart2019precog,salzmann2020trajectron++}

Two prevalent ways for fusing condition are early fusion and late fusion. Early fusion involves encoding condition in the backbone and merging it with the semantic features of the scene. Trajectron++~\cite{salzmann2020trajectron++}, for instance, employs an additional LSTM to encode the future planning of the ego robot and concatenates it with features of surrounding agents for subsequent prediction. SceneTransformer~\cite{ngiam2021scene} can handle multiple conditional tasks within a unified model, by controlling the leakage of future through different temporal masking strategies in its transformer encoder. Late fusion, in contrast, is achieved by fixing rollouts in the iterations of the autoregressive trajectory decoder. To leverage trajectory conditions during decoding, modelling social interaction within the decoder becomes necessary in late fusion. For instance, approaches like ScePT~\cite{chen2022scept}, WIMP~\cite{khandelwal2020if}, MFP~\cite{tang2019multiple} and MotionLM~\cite{Seff2023MotionLM} incorporate an autoregressive trajectory decoder with an agent-attention module during the decoding phase, capturing interaction-aware feature within the future horizon. Moreover, JFP~\cite{luo2023jfp} builds energy-based interaction to approximate joint distribution of all agents, which allows easily performing conditional inference. Specifically, Tolstaya \textit{et al}~\shortcite{tolstaya2021identifying} reported that early fusion outperforms the late fusion in $L_{2}$ metric.

Seff \textit{et al}~\shortcite{Seff2023MotionLM} mentions the concept of temporal causality within the context of CMP. It denotes that predictions about the future at a given moment are contingent exclusively upon conditions leading up to that moment. As for late fusion, approaches with autoregressive decoder based on RNN or GRU are naturally satisfying temporal causality in sequential iteration when achieving CMP~\cite{chen2022scept,khandelwal2020if}. In~\cite{Seff2023MotionLM}, ignoring temporal causality in conditioning improves prediction performance by exposing more information, but may also result in non-causal vehicle reactions.
\section{Method}

This section elucidates our approach, commencing with an exposition of the formulated MPP alongside CMP. Subsequently, we provide a detailed overview of the architecture inherent to our neural network, proficient in multi-agent prediction and planning. Following this, we expound upon the closed-loop planning framework specifically tailored for our neural network, emphasizing its adaptive scheduling capabilities and CMP safety monitoring. Concluding this section, training loss applied to our imitation-based neural planner is introduced.
\begin{figure*}[t]
      \centering
      \includegraphics[width=\linewidth]{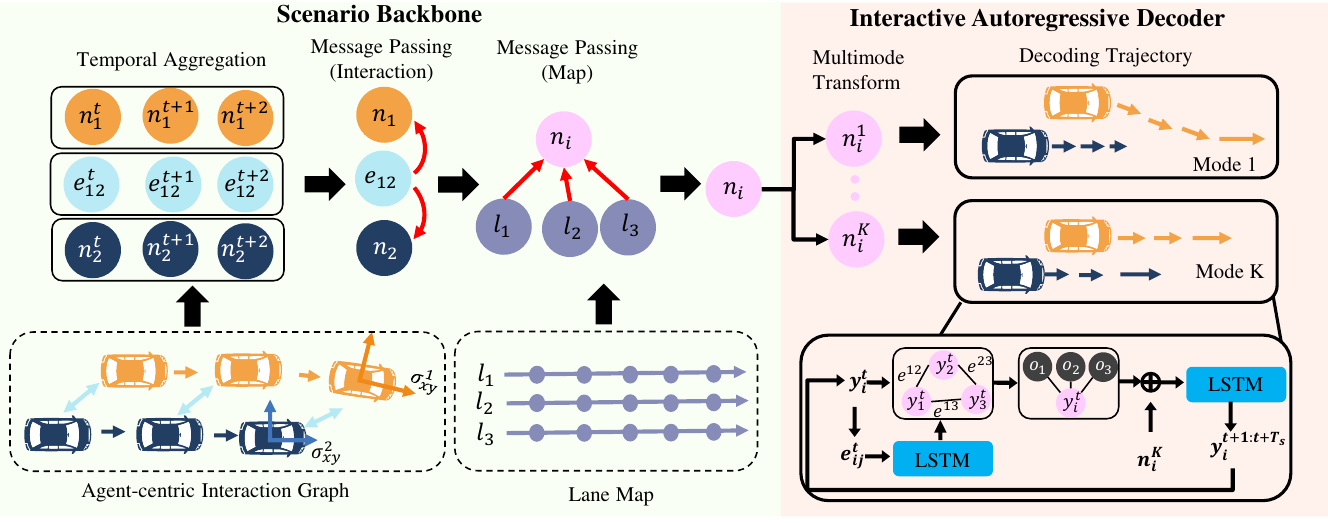}
      \caption{The neural network architecture comprises the scenario backbone and the interactive autoregressive decoder.}
      \label{fig:model architecture}
\end{figure*}
\subsection{Problem Formulation}\label{subsec:problem formulation}
A typical driving scenario consists of an Ego Vehicle (EV), an indeterminate number of Surrounding Vehicles (SV) and Obstacles (O), and Lanes Segments (LS) in static map and various traffic semantics. 
To determine local planning in the future $T$ time, the neural network takes historical trajectories of $ S_{EV} $ and $S_{SV}$ as inputs, producing outputs for the EV planning $Y_{EV}$ and SV predictions $Y_{SV}$. In decoding phase, the following factorization of distribution over joint future trajecgtory is assumed: 
\begin{align}
            P_{\theta}(Y|C_{\eta}) &=  \prod_{\tau=0}^{T} P_{\theta}(Y^{\tau}|Y^{<\tau},C_{\eta}),\\
      = \prod_{\tau=0}^{T} &\prod_{n=1}^{N} P_{\theta}(y^{\tau}_{n}|Y^{<\tau},C_{\eta}), \label{eq: joint prediction and planning}
\end{align}%
where $C_{\eta}$ is context feature from scenario backbone. Since conditioning can facilitate the prediction of SVs~\cite{rhinehart2019precog}, CMP is preferred after ego planning is determined. 
The early fusion of CMP can be described as follows:
\begin{equation}\label{eq:early CMP}
    \resizebox{.91\linewidth}{!}{$
    \displaystyle
    P_{\theta}(Y_{SV}|C_{\eta}(Y_{EV})) = \prod_{\tau=1}^{T} P_{\theta}(y^{\tau}_{SV}|Y^{<\tau}_{SV},C_{\eta}(Y_{EV}=y_{EV})),
    $}
\end{equation}
where the output of scenario backbone $C_{\eta}$ is dependent on the total future $Y_{EV}$.
For comparison, the late fusion in autoregressive decoder can be described as following:

\begin{equation}\label{eq:late CMP}
    \resizebox{.91\linewidth}{!}{$
    \displaystyle
    P_{\theta}(Y_{SV}|Y_{EV},C_{\eta})=\prod_{\tau=1}^{T} P_{\theta}(Y^{\tau}_{SV}|Y^{<\tau}_{SV},Y^{<\tau}_{EV}=y_{EV}^{<\tau},C_{\eta}),
    $}
\end{equation}
where the $Y_{SV}^{\tau}$ is only conditioned on $Y_{AV}^{<\tau}$, satisfying temporal causality~\cite{Seff2023MotionLM}. 
\subsection{Achitecture of Neural Network}\label{subsec:Architecture}
Fig.~\ref{fig:model architecture} depicts the architecture of our neural network. The scenario backbone is a graph model that fuses multimodal environmental context and the trajectory decoder is an interactive autoregressor that aggregates local information in each iterative step. Each agent, along with its relationships, is encoded in their local reference frame denoted as $\sigma_{xy}^{i}$, contributing to a consistent and standardized representation. Similar design choices have been adopted in HiVT~\cite{zhou2022hivt}, wherein the homogeneous agent-centric representation for each agent improves data efficiency. In our network, we not only adhere to this practice in the backbone but also extend it to the autoregressive decoder, enhancing the learning efficiency of agent interaction, improving performance of late fusion for CMP. Due to the homogenous agent-centric representation in the local reference frame, EV and SVs are not distinguished in the backbone unless explicitly mentioned.
\subsubsection{Scenario Backbone.}

\paragraph{Extracting agent-centric interaction.}The scenario is represented as a graph where nodes correspond to agents (vehicles), and edges represent relative relationships among agents or between agents and the environment. For the $i^{th}$ vehicle, positional information $p_{i}^{t} $, motion displacement $\Delta p_{i}^{t}$ and vehicle attributes are encoded:
\begin{equation}
    n_{i}^{A,t} = MLP([R^{T}_{i}(x_{i}^{t}-x_{i}^{0}),R^{T}_{i}\Delta x_{i}^{t},a^{A}_{i}]),
\end{equation}
where $a^{A}_{i}$ is agent-related attributes like length, width and $R_{i}$ is the rotation matrix through current vehicle heading $\alpha_{i}^{0}$. Edge features $e_{ij}^{AA,t}$ capture interactions between nearby agents:
\begin{equation}
    e_{ij}^{AA,t} = MLP([R^{T}_{i}\Delta x_{j}^{t},R^{T}_{i}(x_{j}^{t}-x_{i}^{t})]).
\end{equation}

\paragraph{Temporal aggregation.}To captures the dynamics of motion and interaction, temporal transformers~\cite{zhou2022hivt} are employed. These transformers aggregate node and edge features in the temporal dimension, producing outputs denoted as $n_{i}^{A}$ and $e^{AA}_{ij}$. This process enables the model to effectively capture the evolving patterns and relationships, enhancing its understanding of both the dynamic interactions and motions. 
\begin{figure*}[t]
      \centering
      \includegraphics[width=\linewidth]{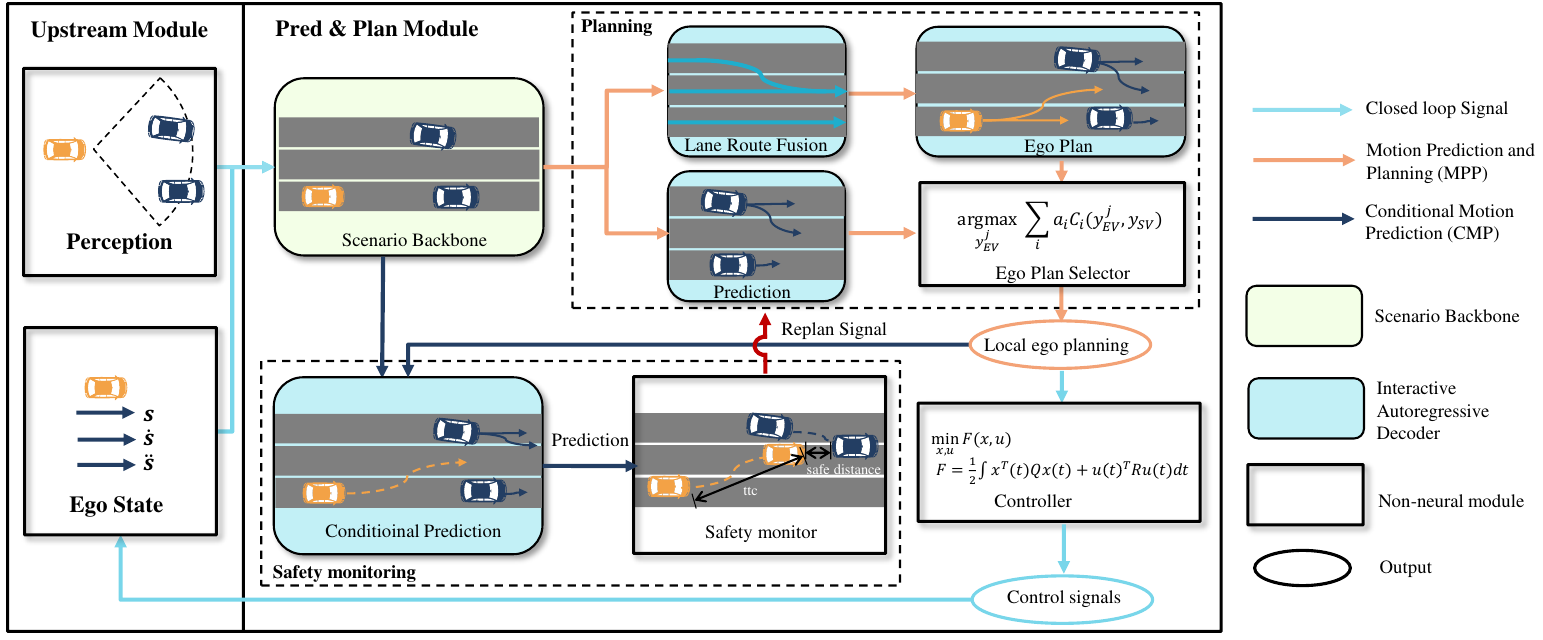}
      \caption{A closed-loop planning framework for our neural network, featuring a planning mode driven by Motion Prediction and Planning (MPP) and a safety monitoring mode with Conditional Motion Prediction (CMP).}
      \label{fig:closed loop framework}
\end{figure*}
\paragraph{Message passing.} Thereafter, the graph attention mechanism~\cite{vaswani2017attention} is employed to aggregate the dynamics of interaction into agent features:
\begin{equation}
    q_{i}=W^{Q}n_{i},\ k_{ij}=W^{K}e^{AA}_{ij},\ v_{ij}=W^{V}e^{AA}_{ij},
\end{equation}
where $W^{Q}$,$W^{K}$,$W^{V}$ are learnable matrices. 
Parallel to the edge features of agent-to-agent interactions, the edge features of lane-to-agent interactions are constructed based on the relative states between the lane and the agent:
\begin{equation}
    e_{ij}^{LA} = MLP([R^{T}_{i}\Delta x^{L}_{j},R^{T}_{i}(x^{L}_{j}-x_{i}^{0}),a_{i}^{L}]),
\end{equation}
where $a_{i}^{L}$ represents attributes of the lane points, such as the status of traffic lights, the type of lane or whether the lane is part of the route. Consistent with the agent-to-agent interaction, map fusion involves an attention mechanism. Finally, self-attention is applied, followed by multimodal projection, to update the agent node feature.

\subsubsection{Interactive Autogressive Decoder.}
Utilizing an autoregressive decoder for decoding future trajectories of agents offers two distinct advantages. Firstly, the iterative nature of the decoder allows for local fusion of scene information with existing trajectories. This temporal iteration facilitates the incorporation of more precise local information and better captures intricate agent interactions~\cite{chen2022scept}. Secondly, the autoregressive approach enables the late fusion of CMP by fixing desired rollouts. Notably, this method naturally adheres to temporal causality, as discussed in Sec.~\ref{subsection:conditional motion prediction}. The ability to extract scene details and maintain temporal coherence makes autoregressive decoding a robust choice for trajectory prediction in dynamic environments.

\paragraph{Iterative interaction.} In order to match the features encoded in the local frame $\sigma_{xy}^{i}$ in backbone, the coordinates $\hat{y}_{i}$ decoded from the decoder are trained in $\sigma_{xy}^{i}$. Therefore, when encoding the interactions from $j$ to $i$, the coordinates $\hat{p}_{j}$ in $\sigma_{xy}^{j}$ need to be restored in frame $\sigma_{xy}^{i}$, which is,
\begin{align}
    \hat{y}_{ij}^{t} = R_{i}^{T}(R_{j}^{-T}(\hat{y}_{j}^{t}+y_{j}^{0})-y_{i}^{0}),\\
    \Delta\hat{y}_{ij}^{t} = R_{i}^{T}(R_{j}^{-T}(\hat{y}_{j}^{t}-\hat{y}_{j}^{t-1})).
\end{align}
The edge features in autoregressive decoder are formulated in the following way:
\begin{equation}
\hat{e}_{ij}^{AA,t} = MLP([\Delta \hat{y}_{ij}^{t},\hat{y}_{ij}^{t}-\hat{y}_{i}^{t}]).
\end{equation}

\paragraph{Local fusion.}In the local fusion process, semantic information is acquired within the point-based graph, which is similar to the approach taken in the agent-lane fusion within the backbone. Also, this involves transforming the coordinates of semantic elements into the corresponding $\sigma_{xy}$. For SV, the fusion process is restricted to incorporating the positions and types of static obstacles. In contrast, for EV, additional local information concerning route lanes is also fused for navigation purpose.

Before feeding the final decoder, the node features of backbone are concatenated together with the current decoding coordinates, and fusion features.
\begin{equation}
    f_{i} = MLP([f_{i},n_{i},s_{i}^{t}]).
\end{equation}


It is important to note that in the decoder, parameters are not shared between EV and SVs, with the sole exception of the interaction module. This design choice facilitates the learning of dynamic interactions between EV and SVs within shared interaction module, while the inductive bias of EV can be fully incorperated from navigation signal or route lane.
\subsection{Closed-loop Planning Framework.}\label{subsec:planning framework}
\subsubsection{Planning and Safety Monitoring}

\begin{figure}[t]
      \centering
      \includegraphics[width=\linewidth]{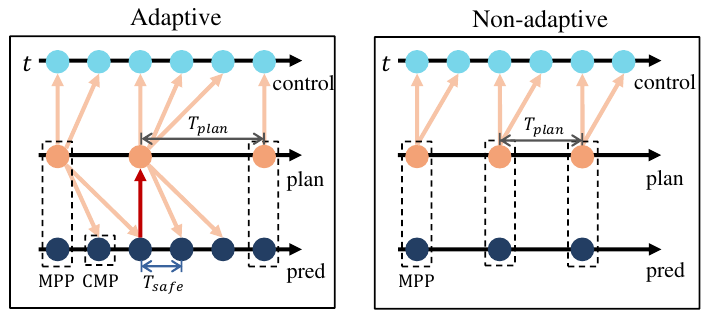}
      \caption{Temporal workflow involves adaptive and non-adaptive scheduling. In non-adaptive scheduling, the replanning time interval $T_{plan}$ remains fixed, requiring a high frequency of MPP to ensure safety. Conversely, adaptive scheduling decouples planning driven by MPP and safety monitor with CMP. This allows for a longer $T_{plan}$ while still ensuring safety.}
      \label{fig:adaptive scheduling}
\end{figure}

Fig.~\ref{fig:closed loop framework} portrays a closed-loop planning framework, leveraging a neural network to integrate EV planning with SV predictions. This planning framework operates in two distinct modes: planning mode and safety monitoring mode. With respect to neural planner, the Motion Prediction and Planning (MPP) and the Conditional Motion Prediction (CMP) are performed correpondingly in each mode. For simplicity, the standard cycle of planning mode is denoted as $T_{plan}$ while that of safety monitoring is denoted as $T_{safe}$. As delineated in Fig.~\ref{fig:adaptive scheduling}, these modes operate at different frequencies, with the planning mode characterized by a relatively low frequency, and the safety monitoring mode featuring a higher frequency, which is: 
\begin{equation}
    T_{plan} = nT_{safe}.
\end{equation}

\paragraph{Planning mode.} In planning mode, operating as a local planner, the neural network jointly generates interactive motion predictions and planning, as outlined in Eq.~\ref{eq: joint prediction and planning}.  The planning mode operates at a low frequency, involving the inference of the scenario backbone and an interactive autoregressor. During decoding phase, the planning head iteratively extracts local information of the lane route for precise driving guidance. In the pursuit of enhanced performance, we employ hybrid metrics to guide the trajectory selection process. Prioritizing safety, our approach adopts a strategy similarly in~\cite{pini2023safe}, which not only considers predicted confidence but also introduces a penalty for time-to-collision:
\begin{equation}
    \hat{p}_{EV,k}=p_{EV,k} - \alpha \mathbb{I}(\min_{i}{\tau_{i,k}}<\tau_{0}).
\end{equation}
Here $p_{k}$ is the predicted confidence for planning $y_{SV,k}$, $\alpha$ is fixed penalty term, $\mathbb{I}$ is indicator function, $\tau_{i,k}$ is time-to-collision of $k^{th}$ planning to the most confident prediction of $i^{th}$ agent and $\tau_{0}$ is predetermined threshold. 

\paragraph{Safety monitoring mode.} The CMP provides predictions based on the current local planning for safety monitoring. The safety monitor diligently observes environmental changes at a higher frequency, promptly triggering a replan when potential dangers are detected. This adaptive planning process is essential to ensure the efficacy of the model. The system operates at a higher frequency to generate SV prediction conditioned on EV planning. The scenario encoding is same to that of MPP stage. With the rollouts of the planning decoder fixed, the predetermined trajectories are fed into the encoding of the interactive graph within the iterative decoder. This integration facilitates the late fusion of ego planning, endowing the model with CMP capabilities. It is important to note that such CMP exhibits temporal causality, as described in Eq.~\ref{eq:late CMP}. 
\begin{table*}[htbp]
  \centering
  
  \begin{tabular}
{c@{\hspace{0.20cm}}|c@{\hspace{0.20cm}}c@{\hspace{0.20cm}}c@{\hspace{0.20cm}}|c@{\hspace{0.20cm}}c@{\hspace{0.20cm}}c@{\hspace{0.20cm}}c@{\hspace{0.20cm}}c@{\hspace{0.20cm}}c@{\hspace{0.20cm}}c@{\hspace{0.20cm}}c@{\hspace{0.20cm}}c@{\hspace{0.20cm}}c@{\hspace{0.20cm}}c@{\hspace{0.20cm}}c@{\hspace{0.20cm}}c@{\hspace{0.20cm}}c@{\hspace{0.20cm}}|c@{\hspace{0.20cm}}}
    \toprule
    &                $T_{pred}$&$\alpha$ &PreType&BLV    &ST     &SL         &HMS    &NMV      &LMS &CL  &FLL &STLT&HLA &LT  &RT  &TPD &WPC & Ave.  \\
    \midrule
    \multirow{4}{*}{NR}
    &0.1&0&MPP    &93.0&96.9   &93.4      &91.3   &88.0      &85.4&88.1&\textbf{96.1}&90.4&74.6&70.5&77.3&65.2&67.5&83.0\\
    &2&0&MPP   &\textbf{97.9}&\textbf{97.4}&\textbf{98.6}&\textbf{93.7}&\underline{93.9}&86.0&\textbf{93.9}&\underline{89.6}&91.6&\textbf{84.6}&\textbf{83.0}&80.1&75.0&80.0&\underline{88.4}\\
    &2&10&MPP   &90.1&93.0&92.6&\underline{93.3}&90.3&\textbf{93.5}&90.2&87.0&\textbf{92.6}&78.2&81.9&\textbf{82.4}&\textbf{78.7}&\textbf{85.1}&87.7\\
    &2&10&CMP&\underline{96.6}&\underline{95.1}&\underline{97.2}&93.1&\textbf{94.1}&\underline{92.3}&\underline{93.5}&87.2&\underline{92.3}&\underline{83.9}&\underline{82.1}&\underline{82.0}&\underline{77.9}&\underline{84.2}&\textbf{88.9}\\
    \midrule
    \multirow{4}{*}{R}
    &0.1&0&MPP&\textbf{92.8}&91.6&\underline{94.0}&86.5&\underline{82.4}      &80.1&81.4&\underline{86.0}&81.4&74.0&60.5&72.9&64.4&69.8&78.7\\
    &2&0&MPP&\textbf{92.8}&\underline{93.7}&\textbf{97.6}&\textbf{90.5}&\textbf{83.2}   &82.8&\textbf{88.4}&\textbf{88.8}&82.2&\textbf{81.3}&72.6&77.8&73.6&78.5&83.7\\
    &2&10&MPP&88.4&\textbf{95.3}&92.8&87.8&79.3&\textbf{86.3}&87.4&73.0&\textbf{86.9}&77.9&\textbf{75.2}&\underline{81.5}&\textbf{79.5}&\textbf{81.7}&\textbf{84.3}\\
    &2&10&CMP&\underline{90.2}&92.5&93.7&\underline{88.1}&80.3&\underline{86.2}&\underline{87.8}&75.5&\underline{84.8}&\underline{78.1}&\underline{74.5}&\textbf{82.7}&\underline{78.4}&\underline{80.5}&\underline{84.0}\\
    \bottomrule
  \end{tabular}
  \caption{Composite scores of each scenario category in \textit{Val14} benchmark, where \textbf{bold} indicates best score and \underline{underline} indicates second-best score. \textit{BLV}:Behand Long Vehicle. \textit{ST}:Stationary in Traffic. \textit{SL}: Stopping with Lead. \textit{HMS}:High Magnitude Speed. \textit{NMV}:Near Multiple Vehicles. \textit{LMS}: Low Magnitude Speed. \textit{CL}: Change Lane. \textit{FLL}:Follow Lane with Lead.   \textit{STLT}: Straight Traffic Light Trasversing.   \textit{HLA}: High Lateral Speeds.  \textit{LT}: Left Turn. \textit{RT}: Right Turn. \textit{TPD}: Trasversing Pickup Dropoff. \textit{WPC}: Waiting for Pedestrian to Cross.}
  \label{tab:ablation study}
\end{table*}

\subsubsection{Adaptive Scheduling}
Compared to rule-based planner, imitation-based planners are susceptible to the inertia problem \cite{codevilla2019exploring}, making them more prone to experience lateral offset overshoot. In Sec.\ref{sec:experiment}, it can be seen that lateral offset overshoot leads to failure by driving off road.

The motivation behind addressing this issue is rooted in striving to complete plans within an extended planning cycle, thereby reducing the occurrence of $G^{2}$ discontinuity between consecutive plans\cite{jian2020multi}. The framework has been intricately designed to respond to emergencies in the environment. Specifically, the MPP mode is activated either when replanning is triggered by the safety monitor with CMP or when a low-frequency planning cycle concludes. The encoder-decoder architecture streamlines this process for great convenience. This adaptive approach ensures that the planning process remains responsive to environmental emergency, while avoiding the limitations associated with a fixed-frequency framework.
\subsection{Loss Function.}\label{subsec:loss function}
In this subsection, $\bar{y}$ is network output and $y$ is the expert demonstration. To improve multimodal trajectory learning, a selective approach is employed where only specific modes of trajectories are chosen for backpropagation:
\begin{equation}
    \mathcal{L}_{reg} = -\frac{1}{NT}\sum_{t=1}^{T}\sum_{i=1}^{N}log\rm{P}(\bar{y}_{i,k^{*}}^{t}-R_{i}^{T}(y_{i}^{t}-y_{i}^{0}))
\end{equation}
where $\bar{y}_{k^{*}}$ is the chosen trajectory and $\rm{P}(\cdot)$ is Laplace distribution. Two selection methods are employed: the first involves choosing the trajectory provided by the network that is closest to the ground truth at time $T$, while the second picks the trajectory with the highest confidence. This dual-selection strategy enhances model diversity and minimizes the gap between training and inference. 

Confidence $p_{k}$, regularized by softmax, is predicted for each mode $k$ of trajectories and the soft label $q_{k}$ is based on final displacement:
\begin{equation}
\begin{aligned}
    \mathcal{L}_{conf}& = -\frac{1}{KN}\sum_{i=1}^{N}\sum_{k=1}^{K}q_{i,k}log(p_{i,k}),\\
    \rm{where}&\ q_{i,k}={softmax}_{k}(-a{||\bar{y}_{i,k}^{T}-y_{i}^{T}||}_{1}).
\end{aligned}
\end{equation}

To enhance obstacle-awareness, obstacle collision penalty is incorporated:
\begin{equation}
    \mathcal{L}_{obs} = \frac{1}{NT}\sum_{t=1}^{T}\sum_{i=1}^{N}\sum_{j=1}^{N_{obs}}log\rm{P}(\bar{y}_{i}^{t}-R_{i}^{T}(o_{j}-y_{i}^{0})).
\end{equation}

The final loss consists of weighted terms described above:
\begin{equation}
    \mathcal{L} = \mathcal{L}_{reg} + \alpha_{1}\mathcal{L}_{conf} +\alpha_{2}\mathcal{L}_{obs}.
\end{equation}

\section{Experiments and Analysis}\label{sec:experiment}
\subsection{Experiment Setup}
Experiments are carried out utilizing the nuPlan dataset \cite{caesar2021nuplan} and its corresponding simulator, including approximately 1300 hours of driving data from diverse locations such as Las Vegas, Boston, Pittsburgh, and Singapore. The dataset features a high level of complexity with various driving scenarios. During the training phase, 1,500,000 scenarios were randomly sampled from the training and validation sets. To ensure diversity, samples are collected at least 1s interval. 

Val14 benchmark \cite{dauner2023parting} is utilized for closed-loop evaluation. This benchmark includes 1,118 scenarios categorized into 14 types. In the Reactive Closed-Loop (R-CL) simulation, surrounding vehicles employ an IDM \cite{treiber2000congested} as latent planner, while in the NonReactive Closed-Loop (NR-CL) simulation, surrounding vehicles are sourced from log replay. \textit{Composite Score} is employed for evaluation, comprising multiple metrics such as whether to collide, whether to drive off road, and progress ratio compared to expert demonstrations etc. \cite{caesar2021nuplan} is refered for more details of composite score. Our experiments are conducted on a workstation with i7-13700k processor and GeForce RTX 3090Ti. For further details of the implementation, please refer to the supplementary materials.
\begin{table}[t]
  \centering
  \begin{tabular}{l|cc}
    \toprule
    Method&NR-CL&R-CL  \\
    \midrule
    GC-PGP${}^{**}$\cite{hallgarten2023prediction} &57.0&54.0 \\
    PlanCNN${}^{**}$\cite{renz2023plant} &73.0 &72.0 \\
    Urban Driver${}^{*}$\cite{scheel2022urban} &47.5 &46.0\\
    PlantTF\cite{cheng2023rethinking}& 84.8&76.8\\
    \midrule
    Ours w/o Adapt. Sched. &84.2 &79.1\\
    Ours w Adapt. Sched. &\textbf{88.9} &\textbf{84.0}\\
    
    \bottomrule
  \end{tabular}
  \caption{Driving scores of learning methods on \textit{Val14}. 
  ${}^{**}$ denotes the results obtained from [Dauner \textit{et al.}, 2023] and ${}^{*}$ is based on implementation of nuPlan codebase. }
  \label{tab:performance comparison}
\end{table}


\subsection{Quatitative Results}
Tab.\ref{tab:ablation study} demonstrates the effects of $T_{plan}$, $\alpha$, and the \textit{Prediction Type} (PreType) in safety monitoring over the composite score in multiple driving scenarios. In Tab.~\ref{tab:performance comparison}, several learning-based approaches are comapred with our method. Tab.\ref{tab:prediction type} presents the prediction and planning results with respect to prediction type of safety monitoring. In Fig.\ref{fig:frequency trends.}, performance trends over $T_{plan}$ both in adaptive and non-adaptive framework are portrayed. 
\subsubsection{Performance Comparison}
Recent proposed PlantTF~\cite{cheng2023rethinking}, as our baseline, is a transformer-based planner, where a RL control adapter is jointly trained in accordance with neural trajectory planner, to bridge the gap between planned trajectory and controlled path. As shown in Tab.~\ref{tab:performance comparison}, our neural planner exhibits comparable results with PlantTF when utilizing a fixed short $T_{plan}$ as in early work. Upon employing our newly designed planning framework with CMP safety monitoring and adaptive long $T_{plan}$, there is a notable performance improvement in both NR-CL and R-CL experiments. 
In our experiments, the parameter $T_{plan}$ is 0.1 s in the old framework and adjusted to 2 s in the new framework. 

\subsubsection{Ablation Studies}
\begin{table}[tbp]
  \centering
  \begin{tabular}{l|ccc}
    \toprule
    PreType&ADE (m)&Collision (\%)& Off-road (\%)  \\
    \midrule
    MPP &0.328 &3.4&2.7 \\
    CMP &0.302&2.9&2.6 \\
    \bottomrule

  \end{tabular}
  \caption{Effects of prediction type in safety monitoring. ADE, obtained from the result of the open-loop evaluation on the validation set, and is the average displacement over the next 5 seconds for the 8 SVs closest to the EV. The rate of Col. and Off-road are from NR-CL experiments.}
  \label{tab:prediction type}
\end{table}
\paragraph{Planning interval $T_{plan}$.}In our framework, $T_{plan}$ is the parameter of paramount significance. Within the simulation environment, the highest attainable frequency is realized when $T_{plan}$ is set to 0.1. Through empirical experimentation, we have discerned that the optimal value for $T_{plan}$ is approximately 2s. In Tab.\ref{tab:ablation study}, the finetuning of $T_{plan}$ proves instrumental in yielding substantial improvements, particularly discernible in turning (LT, RT) or changing lane (CL) scenarios.

\paragraph{Penalty term $\alpha$.} The magnitude of $\alpha$ directly influences the safety penalty associated with the selection of candidate trajectories, with larger values of $\alpha$ imposing a more significant penalty.  An increased $\alpha$ compels the driving strategy to adopt a conservative approach. As evident from the Tab.\ref{tab:ablation study}, opting for a conservative strategy results in higher scores for the planner in scenarios characterized by congestion (LMS, TPD) or a significant presence of pedestrians (WPC), while damaging performance in following leading car (SL, FLL). Such results reveal the inherent tension between optimizing for efficiency and robust safety.

\paragraph{Prediction type.} 
With heavy penalty of time-to-collision, CMP enables the model to achieve the second-highest score in numerous types of scenarios, surpassing MPP as evidenced in Tab.~\ref{tab:ablation study}. This implies that CMP effectively mitigates the over-conservation induced by heavy penalties, striking a balance between efficiency and safety. Furthermore, as shown in Tab.~\ref{tab:prediction type}, CMP contributes to a reduction in the collision rate, attributed to its enhanced predictive capabilities. Notably, the off-road rate remains relatively stable throughout this process.
\begin{figure}[t]
      \centering
      \includegraphics[width=\linewidth]{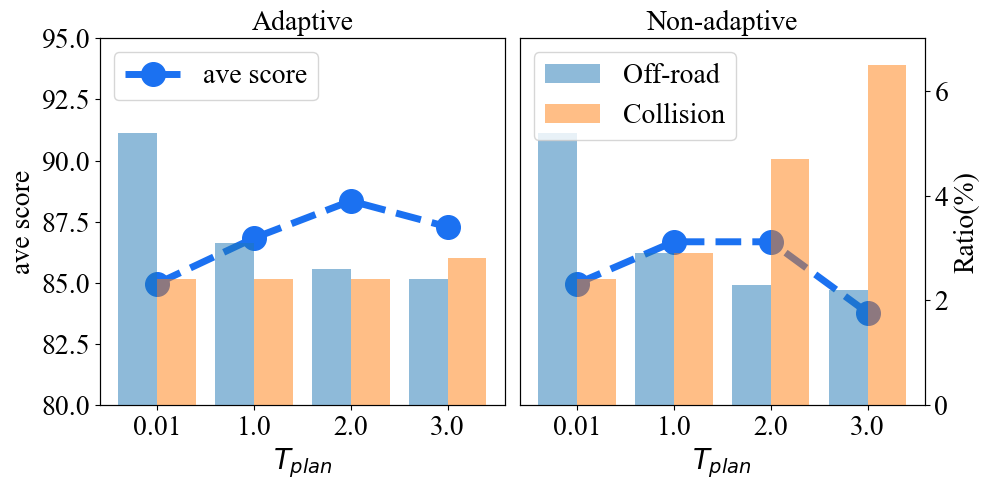}
      \caption{Trends of metrics over $T_{plan}$ for NR-CL simulations in Val14 for adaptive and non-adaptive scheduling. The metrics includes the composite score and ratios of collision and off-road}
      \label{fig:frequency trends.}
\end{figure}
\paragraph{Performance trends over $T_{plan}$.} 
Fig.~\ref{fig:frequency trends.} illustrates the evolving trends of the composite score and the ratio of failure cases with respect to $T_{plan}$. The depicted results showcase our approach alongside those obtained without safety monitoting. Both methods exhibit an initial increase in the composite score followed by a subsequent decrease over $T_{plan}$. This observed trend can be explained by the changes in the ratio of different failure causes. The increase in $T_{plan}$ results in a reduction in the number of vehicles driving off, yet it concurrently leads to an increase in vehicle collisions. Notably, the planning framework equipped with a safety monitor effectively mitigates the trend of escalating vehicle collisions, underscoring its role in suppressing undesirable outcomes. This observation suggests that the inherent inertia problem of imitation-based models might exacerbates lateral offset overshoot over frequent consecutive replanning, ultimately leading to failures in closed-loop simulation, as shown in Sec.~\ref{subsec:qualitative results.}.
\begin{figure}[t]
      \centering
      \includegraphics[width=\linewidth]{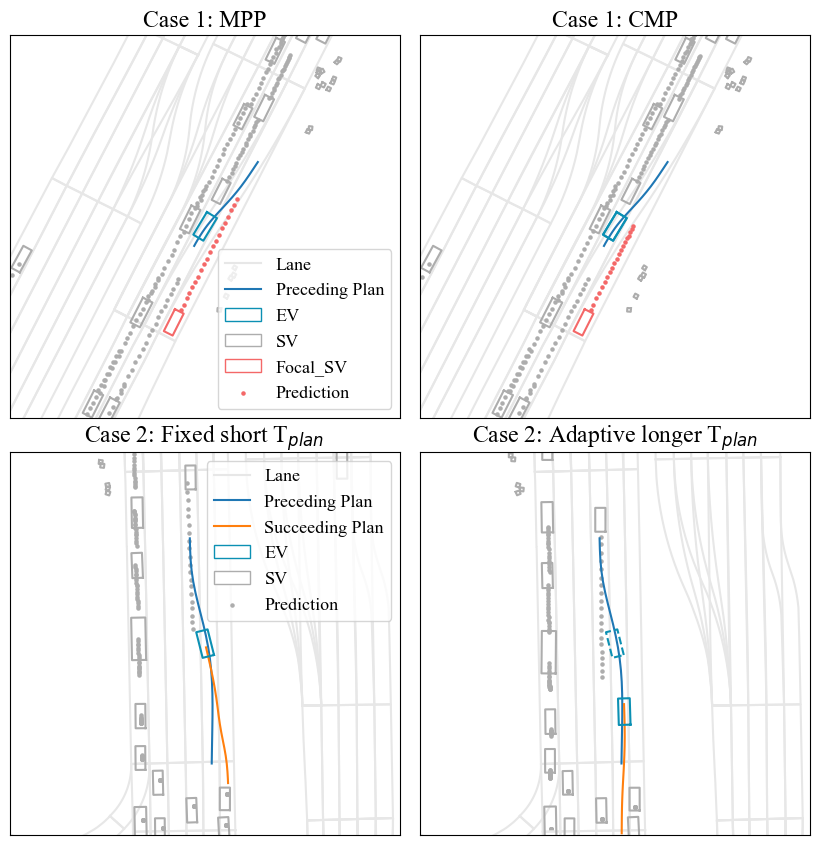}
      \caption{Case study. In Case~1, a comparison is made between the utilization of MPP and CMP in safety monitoring when the Ego Vehicle (EV) plans to change lanes. The MPP predicts a collision between the focal SV and the EV, while the CMP predicts the SV to yield. In Case~2, EV replans when changes lane, leads to unnecessary and continuous lane-changing maneuvers. When applying adaptive longer $T_{plan}$, the EV replans to keep lane.}
      \label{fig:case study}
\end{figure}
\subsection{Qualitative Results.}\label{subsec:qualitative results.}
Qualitative results are shown in Fig.~\ref{fig:case study}. In Case 1, the MPP module anticipates the focal SV traveling normally, predicting an intersection with the lane-changing path tracked by the EV. Consequently, safety monitoring triggers a replanning to prevent the EV from persisting with the lane change. Conversely, CMP predicts that the focal SV is yielding, enabling the lane change to proceed. This discrepancy arises because, during the safety monitoring with MPP, the iteratively decoded prediction of the SV interacts with the currently rolled-out planned trajectory by the planning head, rather than the prior plan of changing lanes. In Case~2, with fixed short replanning interval $T_{plan}$, the EV is more likely to replan when changing lane. As shown in Fig.~\ref{fig:case study}, the succeeding plan and preceding plan is only approximately in $G^{1}$ continuity, leading to unnecessary and continuous lane-changing maneuvers. Conversely, adaptive longer replanning interval results in appropriate replanning to maintain the current lane.

Case.~1 demonstrates that replacing MMP with CMP during safety monitoring can avoid consertive decision and improves the performance in some scenarios, while Case~2 explains why lowering $T_{plan}$ results in a higher rate of off-road, as shown in Fig.~\ref{fig:frequency trends.}.

\section{Conclusion}
This work introduces a novel closed-loop planning framework for neural network integrating prediction and planning to reduce accumulating errors. Through extensive simulation testing, our framework demonstrates capabilities for monitoring safety while ensuring locally stable and feasible trajectory generation from the learned policy. Moreover, our investigations reveal that the integration of CMP into safety monitoring not only augments prediction accuracy but also enhances the overall efficiency of the model. 

Our work develops framework to surface emergent behaviors from neural networks executed in continual feedback. By auditing for stability and safety during closed-loop evaluation, the proposed approach represents initial progress towards trustworthy and reliable autonomous systems.

\appendix



\bibliographystyle{named}
\bibliography{ijcai24}

\end{document}